\documentclass[letterpaper, 10 pt, conference]{ieeeconf}  

\usepackage[T1]{fontenc}
\usepackage{todonotes}
\usepackage{amsmath}
\usepackage{amssymb}
\usepackage{mathtools}
\usepackage[capitalize,noabbrev]{cleveref}
\usepackage{algorithm}
\usepackage{algpseudocode}

\usepackage{multirow}

\IEEEoverridecommandlockouts                              

\title{\LARGE \bf
Diffusion Bridge Models for 3D Medical Image Translation}

\author{Shaorong Zhang$^{1}$, Tamoghna Chattopadhyay$^{2}$, Sophia I. Thomopoulos$^{2}$, \\ Jose-Luis Ambite$^{3}$,  Paul M. Thompson$^{2}$ and Greg Ver Steeg$^{1}$ 
\thanks{$^{1}$ University of California Riverside, CA, United States}%
\thanks{$^{2}$ Imaging Genetics Center, Mark and Mary Stevens Neuroimaging and Informatics Institute, Keck School of Medicine, 
University of Southern California, Marina del Rey, CA, United States}%
\thanks{$^{3}$ Information Sciences Institute, University of Southern California, Marina del Rey, CA, United States 
}
}

\begin{document}

\maketitle
\thispagestyle{empty}
\pagestyle{empty}

\begin{abstract}

Diffusion tensor imaging (DTI) provides crucial insights into the microstructure of the human brain, but it can be time-consuming to acquire compared to more readily available T1-weighted (T1w) magnetic resonance imaging (MRI). To address this challenge, we propose a diffusion bridge model for 3D brain image translation between T1w MRI and DTI modalities. Our model learns to generate high-quality DTI fractional anisotropy (FA) images from T1w images and vice versa, enabling cross-modality data augmentation and reducing the need for extensive DTI acquisition. We evaluate our approach using perceptual similarity, pixel-level agreement, and distributional consistency metrics, demonstrating strong performance in capturing anatomical structures and preserving information on white matter integrity. The practical utility of the synthetic data is validated through sex classification and Alzheimer's disease classification tasks, where the generated images achieve comparable performance to real data. Our diffusion bridge model offers a promising solution for improving neuroimaging datasets and supporting clinical decision-making, with the potential to significantly impact neuroimaging research and clinical practice.

\end{abstract}

\section{INTRODUCTION}

Deep learning has revolutionized our ability to generate and process complex neuroimaging data, particularly in bridging modalities like T1-weighted MRI (T1w MRI) and diffusion tensor imaging (DTI). While both provide crucial insights into brain structure and pathology, DTI is more time-consuming and expensive to acquire, creating a need for efficient cross-modality translation methods. The challenge of acquiring comprehensive DTI datasets is a significant bottleneck in neuroimaging research and clinical practice due to longer scan times, increased patient discomfort, higher costs, and limited availability. This has motivated the development of deep learning approaches to synthesize DTI data from more readily available T1w images, potentially expanding access to diffusion imaging biomarkers while reducing acquisition burden.

Previous approaches to medical image translation \cite{mcnaughton2023machine, chen2023deep} have relied on Generative Adversarial Networks (GANs) \cite{meharban2024t1w, gong2018pet, huang2020mcmt} and Variational Autoencoders (VAEs) \cite{kapoor2023multiscale, liu2021dual}. Gu et al. \cite{gu2019generating} used CycleGAN to map T1 to FA or MD, and vice versa, demonstrating that synthetic FA images can correct geometric distortions in diffusion MRI. However, GANs often suffer from training instability and mode collapse, while VAEs produce blurred outputs that fail to capture fine-grained microstructural information.
Denoising diffusion probabilistic models (DDPMs) have emerged as a promising approach for generating high-quality images through an iterative denoising process \cite{jiang2024fast, muller2023multimodal}. Most studies have focused on translating T1w images to CT or T2w images. More recently, diffusion bridge models \cite{liu2023i2sb, zhou2023denoising, zhang2024exploring} have been developed for a range of image-to-image translation tasks, including edge-to-handbag conversion, depth-to-RGB translation, super-resolution, deblurring, and JPEG restoration. These models have demonstrated superior performance over conditional diffusion models, primarily because they directly establish a transition between source and target images. This direct mapping reduces the distributional gap compared to conventional diffusion models, which must bridge the larger gap between Gaussian noise and the target distribution.


Synthetic neuroimaging data can be beneficial for downstream predictive modeling tasks, such as Alzheimer's disease classification or sex classification (a common benchmarking task with known ground truth). These tasks typically use 3D Convolutional Neural Networks (CNNs) \cite{bin2022adclassification, chattopadhyay2024classification} or vision transformers \cite{dhinagar2023efficiently} to learn discriminative features from volumetric neuroimaging data. Sex classification serves as a benchmark to evaluate the biological plausibility of synthetic data, ensuring that generated images preserve subtle, distributed sex-related anatomical differences. Alzheimer's disease detection is a clinically relevant application that can benefit from enhanced training datasets, as synthetic data can help mitigate the scarcity of labeled data. We focus on these two downstream tasks to evaluate the synthetic scans generated using the diffusion bridge model.

In this work, we adopt diffusion bridge models for 3D medical image translation. Our contributions are summarized as follows: 
\begin{itemize}
    \item We introduce a diffusion bridge model that explicitly captures the joint evolution of T1-weighted (T1w) and Diffusion Tensor Imaging (DTI) domains, enabling anatomically consistent cross-modality translation while preserving the structural integrity of white matter pathways. 
    \item To evaluate the utility of our synthetic data, we conduct extensive experiments on two key predictive modeling tasks: (i) Sex Classification, to verify the preservation of sex-related anatomical features, and (ii) Alzheimer’s Disease (AD) Classification, demonstrating the potential of synthetic DTI in augmenting clinical datasets for improved disease detection.

\end{itemize}

\section{DATA AND PREPROCESSING}


Diffusion tensor imaging (DTI) is a widely used MRI technique that employs diffusion-weighting to model brain tissue microstructure \emph{in vivo}. The diffusion tensor model approximates local diffusion using a spatially varying tensor, represented as a 3D Gaussian at each voxel. Although simpler than advanced approaches like NODDI \cite{deligianni2016noddi} and MAP-MRI \cite{ozarslan2013mean}, DTI remains popular due to its compatibility with single-shell diffusion MRI (dMRI), which is faster and more practical in clinical settings. DTI is summarized using four scalar metrics: fractional anisotropy (FA) and mean, axial, and radial diffusivity (MD, AxD, and RD). These metrics characterize the shape of the diffusion tensor at each voxel, derived from its three principal eigenvalues indicating water diffusion rates along three principal directions. FA summarizes the directionality of diffusion, calculated from the diffusion tensor eigenvalues using a standard formula.


\begin{equation}
    FA = \sqrt{\frac{1}{2}} \frac{\sqrt{\left(\lambda_1-\lambda_2\right)^2+\left(\lambda_2-\lambda_3\right)^2+\left(\lambda_3-\lambda_1\right)^2}}{\sqrt{\lambda_1^2+{\lambda_2}^2+{\lambda_3}^2}},
\end{equation}

where $\lambda_1$, $\lambda_2$ and $\lambda_3$ are the three principal eigenvalues.

The Alzheimer's Disease Neuroimaging Initiative (ADNI) \cite{veitch2019understanding} is a comprehensive, multisite study initiated in 2004, at 58 locations across North America. It aims to collect and analyze neuroimaging, clinical, and genetic data to identify and better understand biomarkers associated with healthy aging and AD. We used data from a total of 1,114 participants (age: 74.36±7.74 years; 562 F/552 M), who had both structural T1w as well as dMRI with a distribution of (592 CN/391 MCI/131 dementia) for our analysis. 


3D T1w brain MRI scans underwent preprocessing steps \cite{chattopadhyay2023predicting, lam20203d}, including N4 bias field correction, skull stripping, registration to a template using 6 degrees of freedom rigid-body registration, and isometric voxel resampling to 2-mm spatial resolution. The resulting pre-processed images were of size 91x109x91 and underwent min-max scaling to range between 0 and 1. This normalization process standardizes image intensity values for subsequent analyses and model training. The preprocessing pipeline ensures that the background of the scans is set to 0 intensity, and due to input normalization before the CNN model, the effect of the original background or intensity range on convolution model performance is negligible. All T1w images were aligned to a common ENIGMA consortium template \cite{jahanshad2013multi}, and dMRI were non-linearly registered to the T1w images. The dMRI processing pipeline is detailed in \cite{chattopadhyay2023predicting, thomopoulos2021diffusion, feng2024deep}.

\section{METHODOLOGY}

\begin{figure*}[t]
    \centering
    \includegraphics[width=0.9\linewidth]{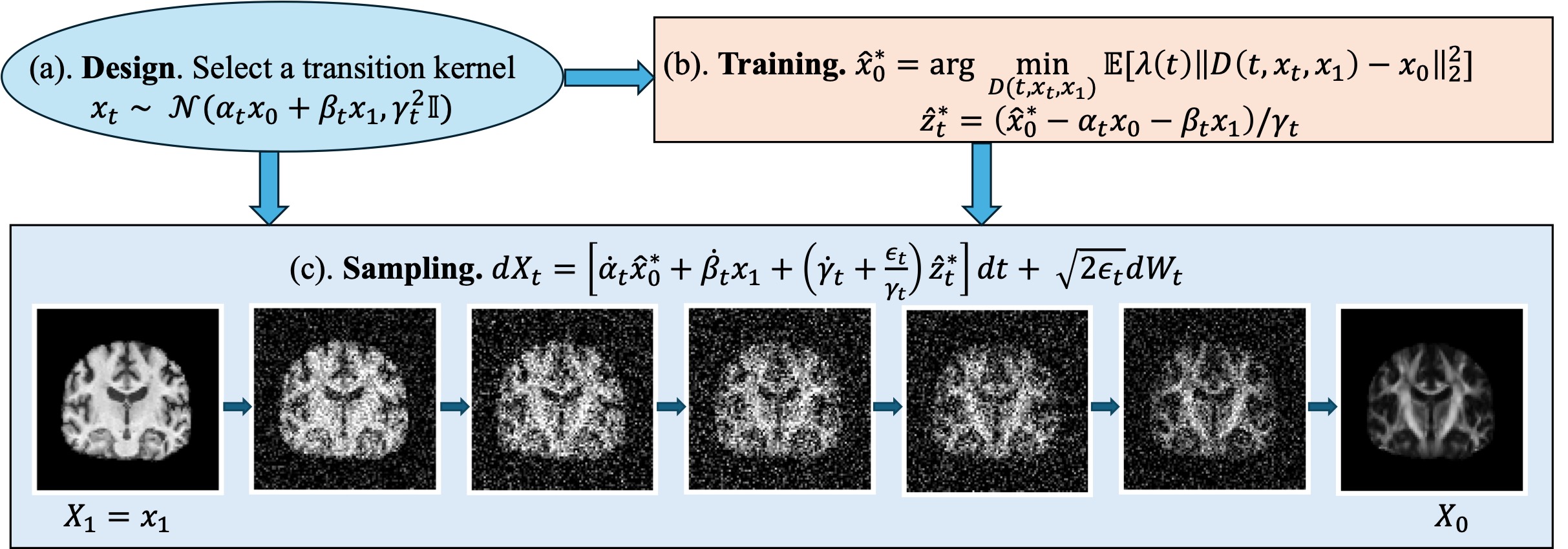}
    \caption{Overall framework of diffusion bridge models for 3D medical image translation.}
    \label{fig:bridge_struc}
\end{figure*}

\subsection{Problem formulation}

In this work, we address the problem of 3D medical image translation.  Given paired source and target images, the goal is to learn a mapping that can predict the corresponding target images for previously unseen source images. 

Let $\{x_1^{i}, x_0^{i} \}_{i=1}^N$ denote a dataset of N paired source and target images, where $x_1^{i}, x_0^i \in \mathbb{R}^d$ and $d$ is the dimensionality of the images. We assume that the source images are drawn from a distribution $\pi_1(x_1)$ and the target images are drawn from a distribution $\pi_0(x_0)$. The joint distribution of source and target images can be represented as $\pi_{0,1}(x_0, x_1) = \pi_1(x_1)\pi_{0\mid1}(x_0\mid x_1)$, where $\pi_{0\mid1}(x_0\mid x_1)$ is the conditional distribution of target image $x_0$ given source image $x_1$.

Our goal is to learn a model of the conditional distribution $\pi_{0\mid1}(x_0 \mid x_1)$ from the paired training data, which can then be used to predict the corresponding target image $x_0$ for a previously unseen source image $x_1$ at test time.





\subsection{Diffusion bridge models} 

Diffusion bridge models \cite{liu2023i2sb, zhang2024exploring, zhou2023denoising} try to learn the conditional distribution $\pi_{0 \mid 1} (x_0 \mid x_1)$, with a method consisting of three steps: designing, training and sampling.  

\subsubsection{Designing}

The first step is to build a stochastic process $p_t$, namely, bridge process, that connects the source $\pi_0$ and target distributions $\pi_1$ and is defined for all $t \in [0, 1]$:

\begin{equation}
\label{eq:bridge_construction}
    p_t = \int p_{t \mid 0, 1} (x_t \mid x_0, x_1) dx_0 dx_1,
\end{equation}

where $p_{t \mid 0, 1}(x_t \mid x_0, x_1)$ is defined as a Gaussian transition kernel:

\begin{equation}
\label{eq:trans_kernel}
    p_{t \mid 0, 1} (x_t\mid x_0, x_1) = \mathcal{N} \left( \alpha_t x_0 + \beta_t x_1, \gamma_t^2 \mathbb{I} \right).
\end{equation}

To ensure the bridge process connects the desired distributions (i.e., $p_1 = \pi_1$ and $p_0 = \pi_0$), the coefficients must satisfy specific boundary conditions \cite{albergo2023stochastic}: $\alpha_t$, $\beta_t$, and $\gamma_t^2$ are differentiable functions of time where $\alpha_0 = \beta_1 = 1$, $\alpha_1 = \beta_0 = \gamma_0 = \gamma_1 = 0$, and $\alpha_t^2 + \beta_t^2 + \gamma_t^2 > 0$ for all $t \in [0, 1]$. 

Intuitively, the transition kernel represents the probability of the system being at $x_t$ at time $t$ given its starting point $x_0$ and endpoint $x_1$; the integral averages over all possible starting and endpoints ($x_0$, $x_1$). Its mean, $\alpha_t x_0 + \beta_t x_1$, is a weighted combination of the starting point $x_0$ and the ending point $x_1$, and the variance is $\gamma_t^2 \mathbb{I}$.
Here $\gamma_t^2$ controls the spread (uncertainty) of the process, and $\mathbb{I}$ is the identity matrix that indicates independence across dimensions. 

The choice of coefficients $\alpha_t$, $\beta_t$, and $\gamma_t$ defines different bridge models, which significantly influences the model's performance. These coefficients determine how the stochastic process interpolates between the source and target distributions, affecting both the learning dynamics and the quality of generated samples. In this work, we adopt a simplified yet effective parameterization recommended in previous studies \cite{zhang2024exploring}:

\begin{equation}
p_{t \mid 0,1}(x_t \mid x_0, x_1) = \mathcal{N}(x_t; (1-t)x_0 + tx_1, 4\gamma_{\max}^2t(1-t)\mathbb{I}).
\end{equation}

This particular choice corresponds to setting $\alpha_t = 1-t$, $\beta_t = t$, and $\gamma_t^2 = 4\gamma_{\max}^2t(1-t)$, which satisfies the required boundary conditions.
Zhang et al. \cite{zhang2024exploring} found that the optimal $\gamma_{\max}$ for image-to-image translation tasks, such as Edges-to-Handbags and Depth-to-RGB image translation, ranges from approximately $0.125$ to $0.25$. In this work, we set $\gamma_{\max} = 0.125$ as the default.

\subsubsection{Sampling}

Given transition kernel $p_{t \mid 0, 1} (x_t \mid x_0, x_1)$ in \cref{eq:trans_kernel}, we can build a bridge process $p_t$ by \cref{eq:bridge_construction}. Then the evolution of conditional probability $p_t(x_t \mid x_1)$ is given by the following Stochastic Differential Equation (SDE):

\begin{equation}
\label{eq:sample_sde}
    d {X}_t = 
    b(t, X_t, x_1) dt + \sqrt{2 \epsilon_t} d {W}_t, \quad X_1 = x_1,
\end{equation}

where $b(t, x_t, x_1) = \dot{\alpha}_t \hat{x}_0^* + \dot{\beta}_t x_1 + (\dot{\gamma}_t + \frac{\epsilon_t}{\gamma_t}) \hat{{z}}_t$, $\hat{x}_0^*(t, x_t, x_1) = \mathbb{E}[x_0 \mid x_t, x_1]$, $\hat{{z}}_t^* =: (x_t - \alpha_t \hat{{x}}_0 - \beta_t x_1) / \gamma_t$. $\epsilon_t> 0$ controls the noise added in the inference time. Notably, $\epsilon_t$ controls the stochasticity added in the inference time, it need to be designed and different $\epsilon_t$ can significantly influence the image translation quality. When $\epsilon_t = 0$, Eq. (\ref{eq:sample_sde}) reduces to a deterministic Ordinary Differential Equation (ODE), while setting $\epsilon_t = \gamma_t \dot{\gamma}_t - \frac{\dot{\alpha}_t}{\alpha_t} \gamma_t^2$ recovers the DDBM sampling SDE \cite{zhou2023denoising}.

\subsubsection{Training}

\cref{eq:sample_sde} is not implementable since $\hat{x}_0^*(t, x_t, x_1)$ is unknown. However, we can train a denoiser $\hat{x}_0^{\theta}$ to approximate $\hat{x}_0^*(t, x_t, x_1)$ via:

\begin{equation}
    \label{eq:objective}
    \hat{{x}}_0^* = \arg \min_{\hat{x}(t, x_t, x_1) }\mathbb{E} \left[\lambda(t) \|\hat{x}(t, x_t, x_1)-{x}_0 \|_2^2 \right],
\end{equation}

where $\lambda(t)$ is a positive weighting function, $\mathbb{E}$ denotes an expectation over $(x_0, x_1) \sim \pi_{0, 1} (x_0, x_1)$, $x_t \sim p_{t \mid 0, 1} (x_t \mid x_0, x_1)$. Although $\hat{{x}}_0^*(t, x_t, x_1)$ can be estimated by a neural network, i.e., $\hat{x}_0^{\theta}(t, x_t, x_1) \approx \hat{{x}}_0^*(t, x_t, x_1)$, in the implementation, we do not directly parameterize $\hat{x}_0^{\theta}$ as a neural network. Instead, we include additional preconditioning steps as in SDB \cite{zhang2024exploring}. 
We include additional pre- and post-processing steps: scaling functions and loss weighting as in SDB \cite{zhang2024exploring}. Let $\hat{x}_0^{\theta}(\mathbf{x}_t, \mathbf{x}_1, t) = c_{\mathrm{skip}}(t) \mathbf{x}_t + c_{\mathrm{out}(t)} (t) F_{\theta} (c_{\mathrm{in}} (t) \mathbf{x}_t, c_{\mathrm{noise}}(t)) $, where $F_{\theta}$ is a neural network with parameter $\theta$, the effective training target with respect to the raw network $F_{\theta}$ is: $\mathbb{E}_{\mathbf{x}_t,\mathbf{x}_0,\mathbf{x}_1,t}\left[\lambda \|c_\mathrm{skip}(\mathbf{x}_t+c_\mathrm{out} {F_\theta}(c_\mathrm{in}\mathbf{x}_t,c_\mathrm{noise})-\mathbf{x}_0\|^2\right]$, 

\begin{align}
    c_{\mathrm{in}}(t) &= \frac{1}{\sqrt{\alpha_t^2 \sigma_0^2 + \beta_t^2 \sigma_1^2 + 2 \alpha_t \beta_t \sigma_{01} + \gamma_t^2}}, \\ c_\mathrm{skip}(t)  &= (\alpha_{t}\sigma_{0}^{2}+\beta_{t}\sigma_{01}) * c_\mathrm{in}^2, \\ c_{\mathrm{out}}(t) &= \sqrt{\beta_{t}^{2}\sigma_{0}^{2}\sigma_{1}^{2}-\beta_{t}^{2}\sigma_{01}^{2}+\gamma_{t}^{2}\sigma_{0}^{2}} c_\mathrm{in}, \\ \lambda &= \frac{1}{c_\mathrm{out}^2}, \quad c_{\mathrm{noise}}(t)=\frac14\log{(t)},
\end{align}

where $\sigma_0^2, \sigma_1^2$, and $\sigma_{01}$ denote the variance of $\mathbf{x}_0$, variance of $\mathbf{x}_T$ and the covariance of the two, respectively.

\begin{algorithm}
\caption{Sampling Algorithm}
\label{algo:diffusion}
\begin{algorithmic}[1]
    \Require model $\hat{x}_0^\theta(x_t, x_1, t)$, time steps $\{t_j\}_{j=0}^N$, input distribution $\pi_{\mathrm{cond}}$, scheduler $\alpha_t, \beta_t, \gamma_t, \eta$
    \State Sample $x_1 \sim \pi_{\mathrm{cond}}$
    \State ${x}_N = x_1$
    \For{$i=N$ \textbf{to} $1$}
        \State $\hat{x}_0 =\hat{x}_0^{\theta} (x_i, x_1, t_i)$
        \State $\hat{z}_i = (x_i - \alpha_{t_i} \hat{x}_0 - \beta_{t_i} x_N)/ \gamma_{t_i}$
        \If{$N \geq 2$}
            \State Sample $\bar{z}_i \sim \mathcal{N} (0, {I})$
            \State $\epsilon_{t_i} = \eta(\gamma_{t_i} \dot{\gamma}_{t_i} - \frac{\dot{\alpha}_{t_i}}{\alpha_{t_i}} \gamma_{t_i}^2$)
            \State $d_i = \dot{\alpha}_{t_i} \hat{x}_0 + \dot{\beta}_{t_i} x_N + (\dot{\gamma}_{t_i} + \epsilon_{t_i} / \gamma_{t_i} ) \hat{z}_i$
            \State $x_{i-1} = x_i + d_i (t_{i} - t_{i-1}) + \sqrt{2 \epsilon_{t_i} (t_{i} - t_{i-1})} \bar{z}_i$
        \Else
            \State $x_{i-1} = \alpha_{t_{i-1}} \hat{x}_0 + \beta_{t_{i-1}} x_N + \gamma_{t_{i-1}} \hat{z}_i$
        \EndIf
    \EndFor
\end{algorithmic}
\end{algorithm}


\subsubsection{Sampling algorithm}
The sampling algorithm design involves two key components: discretization schemes and noise schedule $\epsilon_t$. Based on SDB [1], we implement two discretization approaches:
The first scheme applies Euler discretization:
\begin{align}
\label{discretization_1}
{x}_{t-\Delta t} & \approx {x}_{t} - b(t, x_t, x_1) \Delta t+\sqrt{2 \epsilon_t \Delta t}  \bar{{z}}_t,
\end{align}
where $\bar{{z}}_t \sim \mathcal{N} (0, {I})$.
The second scheme is:
\begin{equation}
\label{discretization_3}
{x}_{t-\Delta t} \approx \alpha_{t - \Delta t} \hat{{x}}_0 + {\beta}_{t - \Delta t} {x}_T  + \tilde{z},
\end{equation}
where $\tilde{z} = \sqrt{\gamma_{t - \Delta t}^2  -  2 \epsilon_t \Delta t}\hat{{z}}_t  +\sqrt{2 \epsilon_t \Delta t}  \bar{{z}}_t$.
We adopt the sampling algorithm from SDB \cite{zhang2024exploring} that integrates both schemes and let $\epsilon_t  = \eta(\gamma_t \dot{\gamma}_t - \frac{\dot{\alpha}_t}{\alpha_t} \gamma_t^2)$. In the implementation we set $\epsilon_t=1$ for stochastic sampling like DDPM \cite{ho2020denoising} and set $\epsilon_t=0$ for determinist sampling like DDIM \cite{song2020denoising}. Algorithm \ref{algo:diffusion} details the complete sampling procedure.

\subsection{3D UNet}

$F_{\theta}$ of the diffusion bridge model is parameterized as ADM's UNet \cite{dhariwal2021diffusion}. ADM has demonstrated remarkable success in image generation \cite{dhariwal2021diffusion}, translation \cite{liu2023i2sb, zhang2024exploring}, and inverse problems \cite{mardani2023variational, song2023pseudoinverse}.   Our implementation extends the original ADM UNet to support 3D medical imaging data while maintaining several key advantages over standard 3D UNet frameworks like MONAI:
1) Flexible architecture supporting customizable attention configurations and dynamic feature scaling;
2) Advanced conditioning through feature concatenation and scale-shift normalization;
3) Modern training optimizations including flash attention, gradient checkpointing, and FP16 precision. The model follows a symmetric UNet structure with input processing, downsampling path, middle attention block, upsampling path with skip connections, and output projection - all adapted for 3D medical imaging data.

\subsection{3D CNN}

We adopt the 3D CNN model to train the classifier for downstream task evaluation, i.e., Sex and AD classification. The 3D CNN model architecture consists of three convolutional blocks, each containing a 3D convolutional layer with a kernel size of 3, instance normalization, max pooling with a kernel size of 2 and stride 2, and ReLU activation. This is followed by a post-convolution block with a 1x1x1 convolution, instance normalization, ReLU activation, and average pooling with a kernel size of 2 and stride 2. Dropout with a rate of 0.3 is applied after the post-convolution block. The final layer is a fully connected layer with 2 output units for binary classification (male/female or AD/non-AD).

\begin{figure*}
    \centering
\includegraphics[width=1.\linewidth]{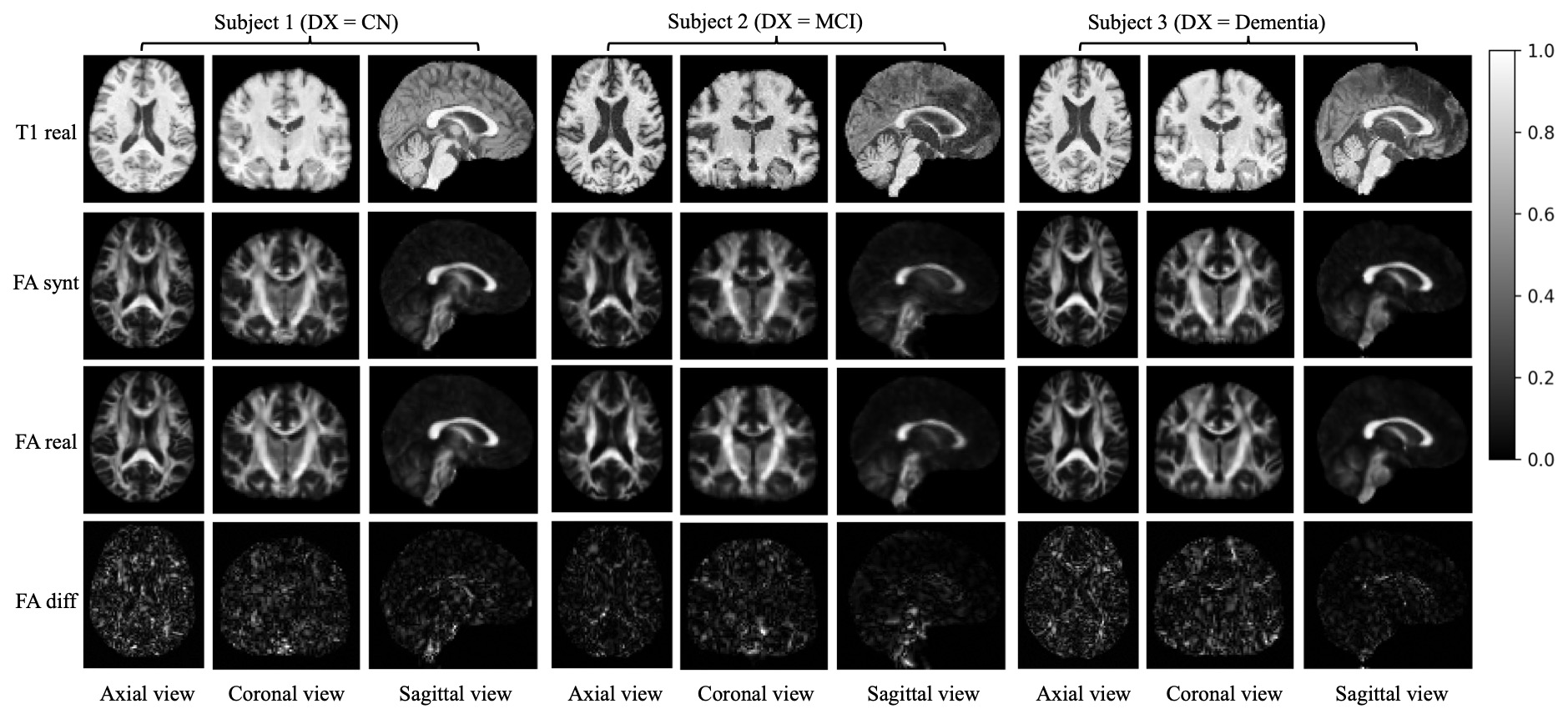}
    \caption{Image translation from T1 to FA with 3 subjects. The table presents four types of images for three different subjects: true T1 images, synthetic FA images, and true FA images. To illustrate the approach, we include example images from participants representing three groups: healthy elderly controls, individuals with mild cognitive impairment (MCI), and those with Alzheimer's disease (dementia).}
    \label{fig:brain_imgs}
\end{figure*}

\begin{figure*}[t]
    \centering
    \includegraphics[width=1.\linewidth]{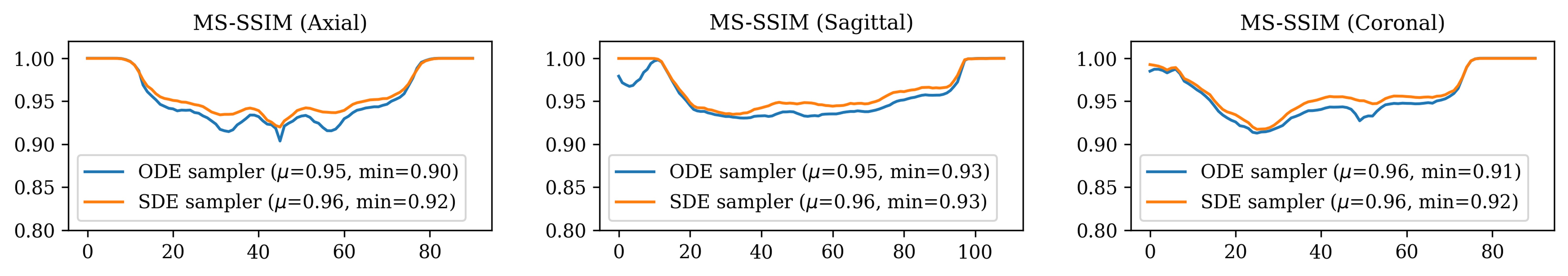}
    \caption{2D MS-SSIM between real and synthetic FA images across different views. $\mu$ represents the mean of MS-SSIM across all slices. The MS-SSIM value for each slice is computed as the mean across 167 subjects in the test dataset. The synthetic FA images are generated by the same pretrained diffusion bridge model with different samplers: ODE sampler ($\eta = 0$), SDE sampler ($\eta = 1.0$).
}
    \label{fig:slices}
\end{figure*}

\begin{figure*}[h]
    \centering
    \includegraphics[width=1.\linewidth]{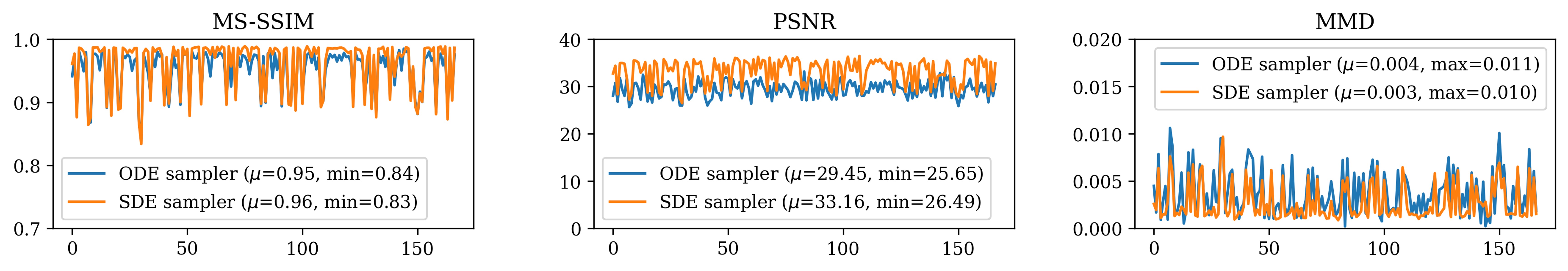}
    \caption{3D MS-SSIM, PSNR and MMD evaluation between real and synthetic FA images across $167$ subjects. The synthetic FA images are generated by the same pretrained diffusion bridge model with different samplers: ODE sampler ($\eta = 0$), SDE sampler ($\eta = 1.0$).}
    \label{fig:metrics_subjects}
\end{figure*}


\section{Experiments}

\subsection{Dataset}

The dataset was randomly split into training, validation, and test subsets in a $7:1.5:1.5$ ratio. This division was used for training the diffusion bridge model as well as for downstream tasks, such as sex and AD Classification. We normalize all brain images to the range $[0,1]$ using min-max normalization. To adapt UNet training, we apply zero-padding to resize brain images from $91 \times 109 \times 91$ to $128 \times 128 \times 128$. This setting is also used for training our Sex and AD classifiers in downstream tasks. In the metric evaluations, we keep the original image size as $91 \times 109 \times 91$.

\begin{table}[h]
    \centering
 \caption{Configurations for diffusion bridge models.}
\label{tab:config_bridge}
\begin{tabular}{ccc}
  \hline
  \textbf{Stage} & \textbf{Hyperparameter}& \textbf{Configuration} \\
\hline
\multirow{1}{*}{Design} & $\gamma_{\max}$ & $0.125$ \\
\hline

\multirow{4}{*}{Train} & learning rate & $5 \times 10^{-5}$ \\
& Batch size & 8 \\
& Micro batch size & 2 \\
& Ema rate & 0.9999 \\
\hline
\multirow{3}{*}{Sampling} & Steps & $40$ \\
&\multirow{2}{*}{$\eta$}& $0$ (ODE sampler)\\
&& $1$ (SDE sampler)\\
\hline
    \end{tabular}
\end{table}
\subsection{Experimental setup}

\subsubsection{Diffusion  bridge model}

Our configurations for diffusion bridge models are presented in \cref{tab:config_bridge}. We adopt the same settings for $\alpha_t$, $\beta_t$, and $\gamma_t$ as in SDB \cite{zhang2024exploring}, with the recommended $\gamma_{\max} = 0.125$. We set the batch size to $8$ and employ a micro-batch processing technique with a micro-batch size of $2$.  In the sampling stage, the number of sampling steps for all experiments is set to $40$ by default. We set $\eta = 1.0$ for stochastic sampling and set $\eta = 0$ for determinist sampling. As $\eta=0$, the diffusion bridge model consistently generates the same image, ensuring reproducibility. 

The architectural specifications for 3D UNet are shown in \cref{tab:hyperparams_unet}. We reduced some hyperparameters to accommodate high pixel ($128\times128\times128$) 3D brain image translation; see \cref{tab:hyperparams_unet} for more details. While reducing architectural parameters (base channels, residual blocks, channel multipliers, attention resolutions) may theoretically limit model capacity, our experiments validate that the streamlined configuration maintains strong performance for medical image translation tasks.

\begin{table}[h]
\centering
\caption{Architectural specifications of our 3D UNet model.}
\label{tab:hyperparams_unet}
\begin{tabular}{ll}
\hline
\textbf{Hyperparameter} & \textbf{Configuration} \\
\hline
Spatial Resolution & $128 \times 128 \times 128$ \\
Input Channels & 1 \\
Base Features & 32 \\
Feature Multipliers & (1×, 2×, 2×) \\
Attention Resolutions & 16, 8 \\
Attention Heads & 1 \\
Head Channels & 16 \\
Dropout Rate & 0.1 \\
Attention Mechanism & Flash \\
\hline
\end{tabular}
\end{table}

\subsubsection{Sex and AD classifier}

The model was trained using the Adam optimizer with a learning rate of 0.0001. The learning rate was adjusted using a ReduceLROnPlateau scheduler, which reduces the learning rate by a factor of 0.5 if the validation loss does not improve for 5 epochs. The model was trained for 50 epochs with a batch size of 8, using cross-entropy loss as the criterion. All experiments used random seed $42$ for reproducibility. 
During training, the best model based on validation accuracy was saved. After training, the best model was loaded and evaluated on the test set to obtain the final test accuracy.
The model's performance was evaluated on both Sex Classification and AD Classification tasks. 

   


\subsection{Metrics}

\textbf{MS-SSIM}. We used the Multi-Scale Structural Similarity Index Measure (MS-SSIM) \cite{wang2003multiscale}, as implemented in the MONAI generative module, to evaluate perceptual similarity between generated and reference images. The MS-SSIM was computed using a Gaussian kernel, and the weights were set to $[0.3, 0.5, 0.2]$ for both 3D and 2D MS-SSIM. 
The MS-SSIM ranges from 0 to 1; higher values indicate greater similarity between the images.

\textbf{PSNR}. We used the Peak Signal-to-Noise Ratio (PSNR) to evaluate the pixel-wise difference between two images. For images normalized to the range $[0, 1]$, the PSNR is calculated as follows:
\begin{equation}
\text{PSNR} = 10 \log _{10} \left(\frac{1}{\text{MSE}}\right),
\end{equation}
where MSE is the mean squared error between the images. Higher PSNR values indicate greater similarity between the images.

\textbf{MMD}. We used the Maximum Mean Discrepancy (MMD) \cite{gretton2012kernel} to quantify the difference between two image distributions based on samples drawn from them. 
We used the implementation provided in the MONAI generative module. Lower MMD values indicate greater similarity between the distributions.


\subsection{Results and analysis}

\subsubsection{MS-SSIM for 2D slices}

Fig.~\ref{fig:brain_imgs} shows that the generated images exhibit crisp anatomical detail that closely resemble the true DTI scalar maps of FA. \cref{fig:slices} shows the 2D MS-SSIM between real and synthetic FA images across different views: Axial view, Sagittal view and Coronal view. The synthetic images exhibit high similarity through the brain, maintaining a score above 0.9 even in anatomically complex interior regions. The main reason is that there is just less brain in the outer images, so the score is higher when more of the image is background. Besides, this may be due to the greater anatomical complexity and density of white matter structures in these central areas.
Additionally, the synthetic FA images generated by the SDE sampler exhibit a slight decrease in MS-SSIM, though the difference is minimal. The ODE sampler, on the other hand, produces deterministic results and offers better reproducibility.

\subsubsection{Metrics for different subjects}

We evaluated MS-SSIM, PSNR, and MMD for $167$ subjects in the test dataset, as shown in \cref{fig:metrics_subjects}. 
Across all three metrics, the vast majority of test synthetic subjects sampled by SDE exhibit high scores, with MS-SSIM approaching 1, PSNR exceeding 30 dB, and MMD remaining below 0.005. These results demonstrate that the proposed method effectively maintains both perceptual fidelity and distributional consistency in the generated images.
However, the presence of a few low-scoring outliers indicates failure cases where the generated images deviate more noticeably from their references. These outliers may stem from limitations in the training dataset, which consists of only 780 image pairs. Expanding the dataset in both size and diversity could potentially enhance the model’s performance on these challenging cases. Additionally, we observe that synthetic images generated using the ODE sampler perform slightly worse in MS-SSIM, PSNR, and MMD evaluations, aligning with the observed trends in 2D MS-SSIM results.

\subsubsection{Downstream tasks}
We evaluated how well the synthetic data retains neurobiologically relevant features via downstream tasks including sex and AD classification (\cref{tab:downstream_tasks}).
Our analysis reveals several key insights regarding the classification performance across different image types. First, for AD classification, the synthetic images exhibit very similar performance to the real images, suggesting that relevant anatomical detail is well-preserved in the bridge diffusion process. However, for sex classification, we note a small but significant drop in performance for the synthetic images, indicating that the translation may be better at preserving some types of anatomical details than others.
Second, an intriguing finding is that synthetic T1 images, although translated from real FA images, achieved better accuracy ($91.7 \%$) than their source FA images ($88.0 \%$) in the AD classification tasks. This suggests that our model not only successfully transfers information between modalities but also enhances certain structural features during the translation process, potentially making them more suitable for classification tasks. Compared to the SDE sampler, the synthetic T1 images generated by the ODE sampler achieve higher accuracy in both AD and sex classification. However, the synthetic FA images exhibit relatively lower accuracy in sex classification. Notably, the ODE sampler produces deterministic images, ensuring reproducibility, while the SDE sampler enhances reconstruction quality in certain tasks.

\begin{table}[h]
    \centering
     \caption{Downstream Tasks Performance: Sex and AD Classification Accuracy.}
    \label{tab:downstream_tasks}
    \begin{tabular}{ccccc}
    \hline
   Training data& Test data & Sampler &AD & Sex\\
    \hline
  \multirow{3}{*}{Real FA} & Real FA& - & 88.0 & 86.8\\
    & Synthetic FA & SDE  & 88.0  & 83.2 \\
     & Synthetic FA & ODE  & 88.0  & 74.3 \\
    \hline
   \multirow{3}{*}{Real T1} & Real T1 & - & 91.7 & 88.6 \\
    & Synthetic T1 & SDE & 90.7 & 82.0 \\
        & Synthetic T1 & ODE & 91.7 & 82.6 \\
         \hline
    \end{tabular}
\end{table}




\section{CONCLUSIONS}

We introduced a diffusion bridge model for 3D brain image translation between T1-weighted MRI and DTI modalities. The model generates high-quality DTI-FA images from T1 images and vice versa, validated using MS-SSIM, PSNR, and MMD metrics. Analysis across brain views and test subjects reveals robustness and generalization, while highlighting potential limitations related to training data size and diversity. Synthetic DTI data retains neurobiologically relevant features, validated through sex and AD classification tasks, showcasing cross-modality information transfer and potential feature enhancement. The approach achieves anatomically consistent translation while preserving white matter pathway integrity, demonstrating its potential for augmenting neuroimaging datasets.

There are several applications of synthetic DTI in research and clinical practice. DTI is not always implemented in resource-limited or time-limited settings, as it is time-consuming to collect and the extra scan time may not be tolerable for some patients. Future work will evaluate the utility of these DTI maps in supporting clinical decision-making. Although the translation works well for patients with MCI and dementia, further work would be needed to determine image translation accuracy when other pathologies are present, such as vascular disease, stroke, or microinfarcts. Ongoing related work in the MRI field includes pan-contrast MRI synthesis \cite{adams2024ultimatesynth} to generate the full range of contrasts attainable with MRI, and MRI harmonization using disentangled latent space representations \cite{zuo2021unsupervised}. Also, diffusion MRI generates a full diffusion function at each voxel, and future work could extend the current model to generate the full tensor or orientation density function, including additional constraints to ensure that the data stays within the relevant manifold (e.g., positive definite tensors and unit mass, symmetric ODFs). Future work will also evaluate performance on larger datasets, higher resolution translation, and refined performance in capturing complex anatomical structures, ultimately impacting neuroimaging research and clinical practice. 

\addtolength{\textheight}{-2.5cm}   

\bibliographystyle{IEEEtran}
\clearpage

\bibliography{refs}




\end{document}